\documentclass[letterpaper, 10 pt, conference]{ieeeconf}  

\IEEEoverridecommandlockouts                              

\title{\LARGE \bf
Switching Attention in Time-Varying Environments via Bayesian Inference of Abstractions
}

\author{Meghan Booker and Anirudha Majumdar$^{1}$
\thanks{$^{1}$ Mechanical and Aerospace Engineering,
        Princeton University, NJ, USA. Emails:
        {\tt\small \{mebooker, ani.majumdar\}@princeton.edu}.
        }
}

\usepackage{amsmath}
\usepackage{amsfonts}

\usepackage{bbm}
\usepackage{url}
\usepackage{graphicx}

\usepackage{algorithm2e}
\RestyleAlgo{ruled}
\usepackage{multirow}
\usepackage{xcolor}

\usepackage{subfigure}

\begin{document}

\maketitle
\thispagestyle{empty}
\pagestyle{empty}

\begin{abstract}
Motivated by the goal of endowing robots with a means for focusing attention in order to operate reliably in complex, uncertain, and time-varying environments, we consider how a robot can (i) determine which portions of its environment to pay attention to at any given point in time, (ii) infer changes in context (e.g., task or environment dynamics), and (iii) switch its attention accordingly. In this work, we tackle these questions by modeling context switches in a time-varying Markov decision process (MDP) framework. We utilize the theory of \emph{bisimulation-based state abstractions} in order to synthesize mechanisms for paying attention to context-relevant information. We then present an algorithm based on Bayesian inference for detecting changes in the robot's context (task or environment dynamics) as it operates online, and use this to trigger switches between different abstraction-based attention mechanisms. Our approach is demonstrated on two examples: (i) an illustrative discrete-state tracking problem, and (ii) a continuous-state tracking problem implemented on a quadrupedal hardware platform. These examples demonstrate the ability of our approach to detect context switches online and robustly ignore task-irrelevant distractors by paying attention to context-relevant information. 

\end{abstract}

\section{Introduction}
\label{sec:intro}

Mechanisms for focusing attention play a pivotal role in human cognition \cite{Fiebelkorn20, Styles06}. In the field of neuroscience and psychology, attention is defined as the \emph{preferential processing} of the stream of sensory information that an agent receives \cite{Styles06, Buschman15}. This preferential processing is crucial for two reasons. First, \emph{bounded rational} agents have a finite computational capacity and must choose certain portions of sensory information to process in order to perform a given task \cite{Simon72}. Second, by focusing only on \emph{task-relevant} information, an agent can achieve robustness to uncertainty and variations in task-irrelevant distractors in its environment \cite{Gigerenzer09, Gigerenzer11, Fiebelkorn20}. The motivating goal of this work is to endow robotic systems with similar capacities for focusing attention.

\begin{figure}[t]
\centering
\includegraphics[width=0.48\textwidth]{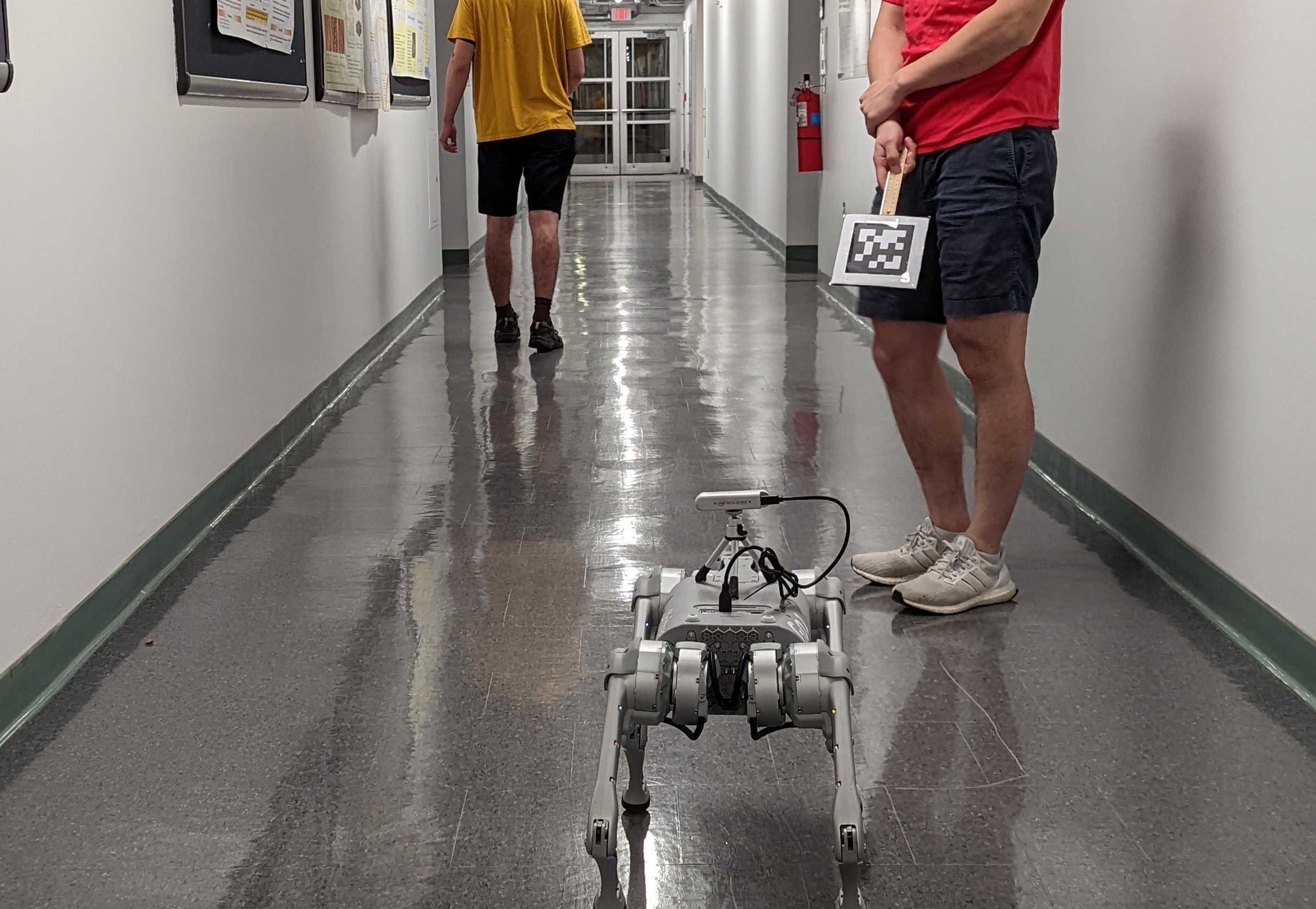}
\caption{A robot dog is tasked with following one of two humans that has a ``treat" (the AprilTag \cite{Olson11} held by the human wearing the red-shirt on the right). The robot observes the state of the two humans (the $x$ and $y$ distances from the robot), infers which human to pay attention to and therefore which to ignore, and then receives the reward of treats if it moves close enough to the human and maintains proximity. The humans can decide to switch who has the treat at any point, and the robot will subsequently need to switch its attention. Note that the robot is not tracking the AprilTag (which may be invisible at any given point in time), but rather the human who is inferred to have the treat.}
\label{fig:hallway_setup}
\end{figure}

Any mechanism for focusing attention must be \emph{context dependent}. In particular, what aspects of a robot's state and environment it should attend to depends on two contextual factors: (i) the task that it is trying to accomplish, and (ii) the dynamics of its environment. As an illustrative example, consider a robot dog tasked with following one of two humans that has a ``treat" (Fig.~\ref{fig:hallway_setup}) that is not always within sight. Intuitively, the robot should pay attention to the state and dynamics of the human that has the treat, and ignore the other human in the environment. Now, if the treat switches from one human to the other, the robot must \emph{infer} this change in context and switch its attention accordingly. Similarly, if the human that does not have the treat changes its behavior and starts interfering with the motion of the human who has the treat, then the robot may need to pay attention to both humans.

In this work, we consider the following question: how can a robot determine what context-relevant information to attend to, infer when the context (i.e., task or dynamics) has changed, and switch its attention accordingly? We hypothesize that these abilities could improve the robustness of robotic systems operating in complex and time-varying environments with multiple sources of potential distraction and uncertainty. 

\textbf{Statement of Contributions.} The main technical contribution of this work is to formalize a means of switching attention to context-relevant information in an online setting. Specifically, we propose a model of context switches in terms of a time-varying Markov decision process (MDP); a given context then corresponds to a particular MDP, and context switches correspond to changes in reward function (i.e., task) and/or dynamics. We leverage the theory of \emph{bisimulation-based state abstractions} \cite{Larsen91, Givan03, Li06} in order to define a mechanism for focusing attention in a given context. We then develop a Bayesian inference approach for determining when the underlying context has switched in order to switch between different 
 abstraction-based attention mechanisms. We demonstrate the efficacy of our approach in terms of automatically inferring context switches and providing robustness to task-irrelevant distractors using two examples: (i) a simulated discrete-state tracking example, and (ii) a continuous-space tracking problem implemented on a quadrupedal hardware platform (Fig.~\ref{fig:hallway_setup}).

\section{Related Work}
\label{sec:related work}

{\bf Bioinspired attention.} 
Mechanisms for focusing and switching attention have been extensively studied in cognitive science (see, e.g., \cite{Fiebelkorn20,Styles06, Buschman15}). As highlighted in Sec.~\ref{sec:intro}, the ability to attend to task-relevant sensory information based on context affords advantages in terms of computational efficiency and robustness to irrelevant distractors.
Inspired by these potential benefits, researchers in control theory and robotics have sought to imbue artificial agents with various forms of bioinspired attention, e.g., attention to visual features \cite{Vijayakumar01, Rea13} or to particular agents in multi-agent settings \cite{Bizyaeva21, Bizyaeva22}. These works rely on hand-crafted attention mechanisms (e.g., via carefully chosen feature representations). In contrast, our goal is to develop a systematic procedure for allowing a robot to decide what to pay attention to.

{\bf Attention via state abstractions.} In this work, we leverage the theory of \emph{state abstractions} \cite{Li06} for Markov decision processes (MDPs). Given a MDP, state abstraction techniques seek to find a mapping from the original state space (which can be large or high dimensional) to a compressed \emph{abstract} state space. This mapping is constructed such that states which differ in a task-irrelevant manner are mapped to the same abstract state; the mapping thus defines a mechanism for focusing attention only on task-relevant information. This intuition is formalized by several state aggregation techniques based on the theory of bisimulation \cite{Larsen91} and its variants (e.g., bisimulation metrics \cite{Ferns11, Zhang20}, lax bisimulations \cite{Taylor08, Ravindran03, VanDerPol20}, and policy-based bisimulations metrics \cite{Castro20, Agarwal21}). We provide a brief introduction to bisimulation and highlight its relationship to attention in Sec.~\ref{background}. 

Prior work on bisimulation typically assumes that the mapping from original to abstract states is \emph{fixed} (since the agent is assumed to be operating in an environment described by a given MDP); thus, what the agent attends to does not change over time. Motivated by the need for robots to operate in complex and changing environments, our goal is to allow for \emph{context-dependent} attention mechanisms. We formalize this in Sec. \ref{sec:problem formulation} by allowing the underlying environment MDP to change as the robot is operating. The robot must then infer when such switches occur and correspondingly switch what it is paying attention to by choosing an appropriate state abstraction. 


{\bf Learning attention mechanisms end-to-end.} An alternative to bisimulation-based attention is to learn the attention mechanism in an end-to-end manner. Attention mechanisms in transformer architectures \cite{Vaswani17} have demonstrated remarkable empirical success in language \cite{Vaswani17, Brown20} and vision tasks \cite{Dosovitskiy20, Ramesh21}. Inspired by these successes, researchers in reinforcement learning and robotics have sought to end-to-end learn similar attention mechanisms \cite{Tang20, Chen21, James22}. Visual attention mechanisms have also been learned via imitation learning \cite{Pfeiffer22, Makrigiorgos19} (e.g., from human gaze observations). These approaches could feasibly learn context-dependent attention mechanisms automatically. However, sample efficiency remains a significant challenge for such end-to-end learning approaches. In contrast, we pursue a model-based approach in this work. 


\section{Problem Formulation}
\label{sec:problem formulation}
Our goal is to provide a robot with a systematic means of focusing and \textit{switching} its attention online. In other words, the robot needs to be able to determine what abstract state information is relevant, infer when the context it is operating in changes, and then correspondingly switch its attention mechanism (i.e., focus attention on new relevant abstract state information in the new context). To formalize this, we assume that the true underlying environment the robot is operating in is described by a time-varying Markov decision process (TVMDP) \cite{Liu18}, $\mathcal{M}_t = (\mathcal{S}, \mathcal{A}, \mathcal{P}_t, \mathcal{R}_t)$, where $\mathcal{S}$ is the discrete or continuous state space of the robot and its environment, $\mathcal{A}$ is the action space, $\mathcal{P}_t(s_{t+1}|s_t, a_t)$ is the time-varying probability of transitioning from state $s_t$ to $s_{t+1}$ with action $a_t$, and $\mathcal{R}_t(s_t) \in \mathbb{R}$ is the time-varying function that assigns a reward for being in state $s_t$.\footnote{The TVMDP is different from a time-dependent MDP that augments the state with time. The time-dependent MDP is interpreted as modelling stochastic travel between states \cite{Liu18}.} In this environment, the robot chooses actions according to a time-varying policy, $\pi_t(s_t)$, with the goal of maximizing the expected cumulative reward, $\max_{\pi_t} \mathbb{E}_{\mathcal{P}_t} [\sum_{t=0}^T \mathcal{R}_t(s_t)]$. A challenge, then, for determining where to focus attention to lies in the fact that the state transition probabilities and reward functions are time-varying; as a consequence, what the robot should attend to is also time-varying (i.e., the relevant abstract state information is time-varying).

We define a \textit{context} to be a time-invariant MDP, $\mathcal{M}^i = (\mathcal{S}, \mathcal{A}, \mathcal{P}^i, \mathcal{R}^i)$, indexed by $i \in \{1, 2,\dots, N\}$, that captures an instance of the true underlying environment $\mathcal{M}_t$. In other words, we assume that at any time, $t$, the underlying environment is some context, $\mathcal{M}^i$. However, the robot does not have knowledge of the context it is operating in. Instead, it can observe an imperfect version of the state and the reward (e.g., noisy depth measurements). 

Next, we assume that for each context, $\mathcal{M}^i$, we are given an \textit{abstraction}, $\phi^i = (\mathcal{Z}^i, \mathcal{A}, \mathcal{P}^{\mathcal{Z}^i}, \mathcal{R}^{\mathcal{Z}^i})$, defined as a MDP obtained from state abstractions for a context. Here, $\mathcal{Z}^i$ is the compressed abstract state space of $\mathcal{S}$, $\mathcal{P}^{\mathcal{Z}_i} := \mathcal{P}^i(z^i_{t+1}|z^i_{t}, a)$ is the probability of transitioning from abstract state $z^i_t$ to $z^i_{t+1}$, and $\mathcal{R}^{\mathcal{Z}_i}:=\mathcal{R}^i(z^i_t)$ is the reward function for the $i$th abstraction. These abstractions provide a mechanism for focusing the robot's attention to relevant state information. In particular, by constraining the robot's policy to only depend on the \emph{abstract} state, task-irrelevant differences in the full state are ignored by the robot. We provide a brief introduction to the bisimulation-based state abstractions used in this paper in Sec.~\ref{background}.

The robot has a catalog, $\Phi=\{\phi^0, \phi^1, \dots, \phi^N\}$, of possible abstractions available to it and needs to infer which one corresponds to the context it is operating in. By construction, inferring abstraction $\phi^j$ at time $t$ and then $\phi^i$ at $t+1$ corresponds to switching the robot's attention since $\mathcal{Z}^i$ now abstracts the state space $\mathcal{S}$ in a different way from $\mathcal{Z}^j$ in order to capture differing relevant information. For the scope of this work, we assume that the catalog is finite. 

While inferring a particular abstraction, the robot chooses actions $a\in \mathcal{A}$ from a given policy $\pi^i(z^i_t)$, i.e., the robot has a policy maximizing the expected cumulative reward for an abstraction.\footnote{Depending on the state abstraction method, the optimal policy of the context may be preserved \cite{Li06}.} The context may switch online, and thus, the robot needs to infer the correct abstraction in order to perform well in the true context. This is the primary objective of this work as this corresponds to determining what the robot is paying attention to and when to switch. We describe our approach for inferring abstractions in Sec. \ref{sec: approach}.

\section{Attention via Bisimulation}
\label{background}

In this section, we provide an overview of bisimulation-based state abstraction \cite{Larsen91, Givan03, Li06} and its relation to attention. Intuitively, bisimulation is a state abstraction method that groups states into the same ``abstract" state if they differ only in a behaviorally irrelevant way. More precisely, bisimulation considers two states to be equivalent if they have (i) equivalent immediate rewards, and (ii) equivalent distributions over the next abstract state. This is formalized in the definition below.

\vspace{3mm}
\noindent
\textbf{Definition 1} (Bisimulation Relation \cite{Givan03}).
\label{def: bisim}
\textit{Given a Markov Decision Process $\mathcal{M}$, a bisimulation relation is an equivalence relation $B$ between states such that for all states $s_i, s_j \in \mathcal{S}$ that are equivalent under $B$, the following conditions hold:
\begin{align}
    &\mathcal{R}(s_i, a) = \mathcal{R}(s_j, a), \quad \forall a \in \mathcal{A} \\
    &\mathcal{P}(z|s_i, a) = \mathcal{P}(z|s_j, a), \quad \forall a \in \mathcal{A}, \ \forall z\in \mathcal{S}_B,
\end{align}
where $\mathcal{S}_B$ is the partition of $\mathcal{S}$ under the relation B (the set of all groupings $z$ of equivalent states), and $\mathcal{P}(z|s, a) := \sum_{s' \in z} \mathcal{P}(s'|s, a)$.}
\vspace{3mm}

The bisimulation relation induces a mapping from the full state $s$ to an \emph{abstract state} $z$ (corresponding to the set of states that are equivalent to $s$ under the bisimulation relation). We can interpret this mapping as a form of attention. Specifically, by mapping two states that are behaviorally equivalent (in the formal sense of Definition 1 to the same abstract state, bisimulation focuses only on relevant differences in the state. 

As an example, consider the problem of following one of two humans with a treat (described in Sec.~\ref{sec:intro} and Fig.~\ref{fig:hallway_setup}). Suppose that \texttt{human1} has the treat. In this case, any state $s$ (corresponding to the locations of the two humans relative to the robot) that differ \emph{only} in terms of \texttt{human2}'s location are equivalent under the bisimulation relation (since \texttt{human2}'s location does not impact the immediate reward or the dynamics of \texttt{human1}'s relative location). Thus, the full state $s$ is mapped to an abstract state (corresponding to \texttt{human1}'s location relative to the robot), which focuses attention on \texttt{human1}. We highlight an important benefit of focusing attention in this manner. Any control policy that depends only on the abstract state has a very strong robustness property: its performance is not impacted by errors in estimation of the irrelevant portions of the state (e.g., \texttt{human2}'s position). 

We note that there are a number of variations of the bisimulation framework discussed above which seek to relax the strictness of the equivalence relation, e.g.,  bisimulation metrics \cite{Ferns11, Zhang20}, lax bisimulations \cite{Taylor08, Ravindran03, VanDerPol20}, and policy-based bisimulations metrics \cite{Castro20, Agarwal21}. In this work, we will utilize the notion of bisimulation in Definition 1. Extending our framework to other notions of bisimulation is a promising direction for future work.

\begin{figure*}[t]
    \centering
    \vspace{2mm}
    \begin{minipage}[t]{0.98\textwidth}
    \centering
    \begin{subfigure}{}
        \centering
        \includegraphics[width=0.31\textwidth]{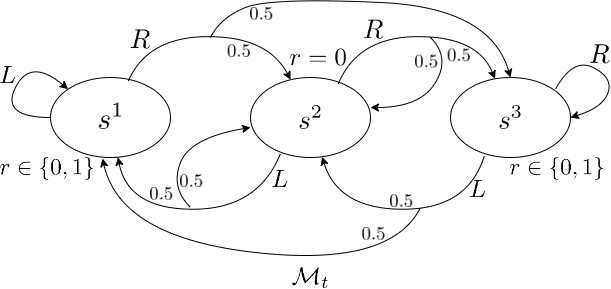}
    \end{subfigure}
    \begin{subfigure}{}
        \centering
        \includegraphics[width=0.31\textwidth]{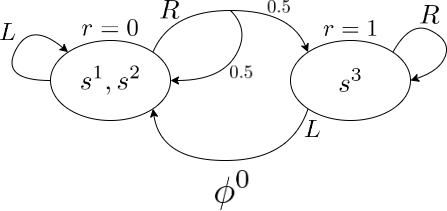} 
    \end{subfigure}
     \begin{subfigure}{}
        \centering
        \includegraphics[width=0.31\textwidth]{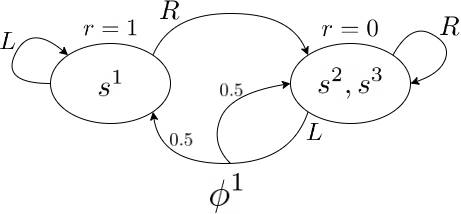} 
    \end{subfigure}
    \end{minipage}
    \vspace{-2.0mm}%
    
    \begin{center}
     \footnotesize(a) \hspace{52mm} (b) \hspace{52mm} (c) 
    \end{center}
    \caption{%
    The problem setup for the discrete tracking example. The robot needs to track the object but the object may appear in states $s^1$ or $s^3$ due to unexpected environment disturbances. When the robot is in the same state as the object, it receives a reward of 1 otherwise 0. \textbf{(a)} The TVMDP for this example. States are denoted in the ovals, actions are either L or R, the state transition probabilities are labelled on the arrows, and the rewards for each state are denoted outside the states with r. \textbf{(b)} The first abstraction, $\phi^0$, acquired by performing bisimulation for when the object is in state $s^3$. States $s^1$ and $s^2$ are mapped to an abstract state together while state $s^3$ is its own abstract state. This corresponds to the robot paying attention to being with the object or not. \textbf{(c)} The second abstraction, $\phi^1$, acquired by performing bisimulation for when the object is in state $s^1$.%
    }
    \label{fig:example1_mdps}
    \vspace{-3.0mm}
\end{figure*}

\section{Switching Attention}
\label{sec: approach}
The robot needs to infer which abstraction, $\phi^i$, to apply online for a given time step in order to pay attention to task-relevant state features and perform well in the true underlying context. To do so, we first use bisimulation, discussed in detail in Sec. \ref{background}, as the state abstraction method for generating the catalog of abstractions, $\Phi$, available to the robot. We use bisimulation because (i) we can leverage the reward as a signal for inferring the correct abstraction to apply (as elucidated further below), (ii) we obtain policies that are robust to noisy state observations, and (iii) there are existing approaches for synthesizing the abstractions and corresponding policies (e.g., \cite{Larsen91, Givan03}).

We now describe our approach for tackling the problem formulated in Sec.~\ref{sec:problem formulation} and inferring which abstraction (and corresponding policy) to apply at any given time. Since the identity of the underlying context is not observed directly, we implement a Bayesian filter to maintain a belief over the abstraction catalog. Specifically, the robot is provided with an abstraction transition model $p(\phi_t^i|\phi_{t-1}^j)$, an abstraction detection model $p(r_t|\phi_t^i, s_t)$, and an initial prior belief over the abstraction catalog $p(\phi^i_0)$. The abstraction transition model captures the uncertainty associated with how the abstractions, and therefore contexts, evolve over time. This and the prior belief provide an opportunity to encode expert or domain knowledge if it is available. The abstraction detection model is pre-generated using the abstractions available to the robot. The probability of a particular reward is one if it matches the abstract state's reward associated with the given abstraction $\phi^i$ and observed state $s_t$. Thus it acts as a ``detector" for an abstraction switch; intuitively, if the robot receives rewards that deviate from its current belief over abstractions, it can infer that the underlying context has switched.  

After updating its belief over abstractions, the robot chooses which abstraction to apply by taking the maximum likelihood estimate. This selection effectively maps the observed, and potentially noisy, state $s_t$ to the abstract state $z^i_t$ and allows the robot to employ the corresponding policy $\pi^i(z_t^i)$. We summarize this attention switching process in Algorithm \ref{alg:bayes} and refer to it as Online Attention Switching (OAS). 

\begin{algorithm}[t]
\caption{Online Attention Switching (OAS)}\label{alg:bayes}
 Initialize environment state: $s_0 \sim \mathcal{S}$ \\
 Initialize maximum likelihood (ML) abstraction: $i \leftarrow \arg \max p(\phi_0^i)$ \\

 \For {$t \in \{1, 2, \dots, T\}$}{
 {Apply action according to ML abstraction}:\\
  $a_t \leftarrow \pi^i(z^i_{t-1})$ \\
 {Robot takes action $a_t$ and observes $s_t, r_t$.} \\
  \For {all $\phi^i$}{
  {Dynamics update:} \\
 $p(\phi_{t}^i | r_{1:t-1}, a_{1:t}, s_{t-1}) = \sum_j p(\phi_t^i|\phi_{t-1}^j)p(\phi_{t-1}^j|r_{1:t-1}, a_{1:t-1}, s_{t-1})$ \\
 {Measurement update:} \\
 $p(\phi_t^i|r_{1:t}, a_{1:t}, s_t) = \frac{p(r_t|\phi_t^i, s_t)p(\phi_t^i | r_{1:t-1}, a_{1:t}, s_{t-1})}{\sum_j p(r_t|\phi_t^j, s_t)p(\phi_t^j | r_{1:t-1}, a_{1:t}, s_{t-1})}$ 
 }
  $i \leftarrow \underset{i}{\text{argmax}} \ p(\phi_t^i|r_{1:t}, a_{1:t}, s_t)$ 
}

 
\end{algorithm}

\section{Examples}
\label{sec:examples}
Here we illustrate the efficacy of our OAS algorithm described in Sec. \ref{sec: approach} on two examples: (i) an illustrative discrete-space tracking example, and (ii) a continuous-space tracking problem where a robot dog has to follow the human with the ``treat". We use these examples to demonstrate that OAS is able infer the correct abstractions efficiently (and therefore perform well in the true contexts) as well as highlight the robustness benefits acquired from using attention. 

\subsection{Discrete-Space Tracking Example}
\label{sec:discrete example}

{\bf Scenario.} In this first example, we consider a robot that has to track an object in a three-state grid-world where the object's location suddenly changes and can appear in either the first or last state. For example, ocean currents or wind may be pushing the object to the different states. The robot can move to follow the object with actions \texttt{left} or \texttt{right} and if it is in the same state as the object, it receives a reward of 1; otherwise it receives a reward of 0. Concretely, the state space is $\mathcal{S}=\{s^1, s^2, s^3\}$, the action space is $\mathcal{A} = \{L, R\}$, the state transition probabilities $\mathcal{P}$, illustrated in Figure \ref{fig:example1_mdps}(a), are time-invariant, and the possible reward functions are $\mathcal{R}_t \in \{ \mathbbm{1}_{s_t=s^1}(s_t), \mathbbm{1}_{s_t=s^3}(s_t)\}  $ where $\mathbbm{1}_{s_t=x}(s_t)$ is 1 if the robot is in state $x$ and 0 otherwise. 

{\bf Contexts and abstractions.} The possible contexts for this TVMDP are induced by the two possible reward functions. Performing bisimulation on these two contexts, as defined in Definition 1, results in the abstractions illustrated in Figure \ref{fig:example1_mdps}(b-c), where the abstract states are capturing whether the robot is paying attention to if it is with the target object or not. Notice that with these abstractions, the robot is robust to noisy state observations. For example, in $\phi^0$, states $s^1$ and $s^2$ are mapped to the same abstract state. Thus, if the true state is $s^2$ but the observed state is $s^1$, the robot will still perform the same actions since $\phi^0$ considers these to be the same state, i.e., the robot is paying attention to the fact that the is not with the object it is tracking not whether it is in exactly $s^1$ or $s^2$.

\begin{figure*}[t]
    \centering
    \vspace{2mm}
    \begin{minipage}[t]{0.99\textwidth}
    \centering
    \begin{subfigure}{}
        \centering
        \includegraphics[width=0.98\textwidth]{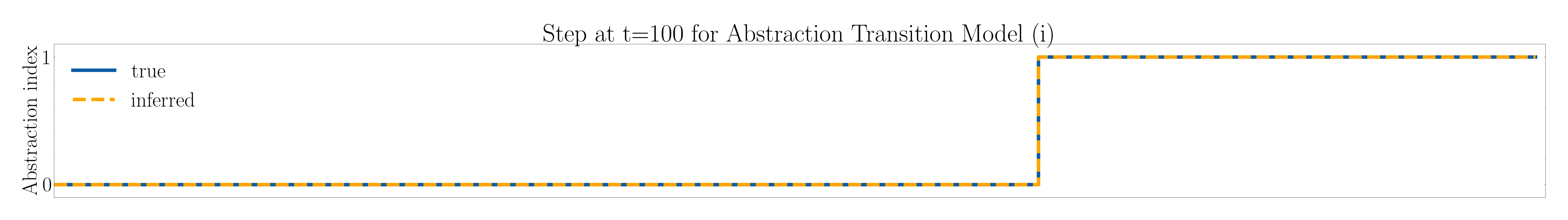} \\
    \end{subfigure}
    \begin{subfigure}{}
        \centering
        \includegraphics[width=0.99\textwidth]{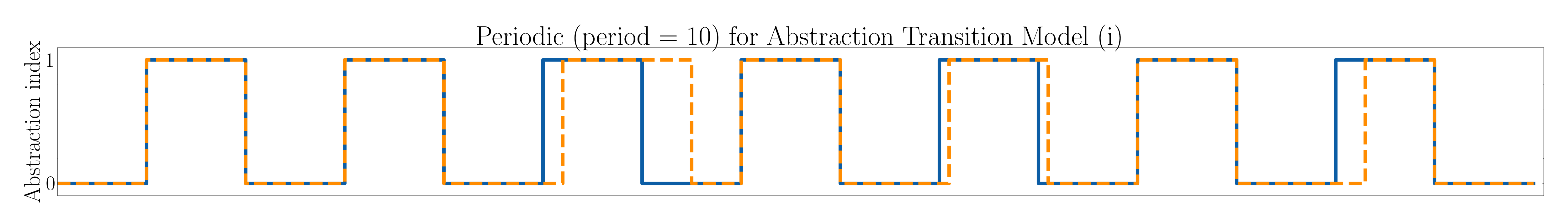} \\
    \end{subfigure}
    \begin{subfigure}{}
        \centering
        \includegraphics[width=0.99\textwidth]{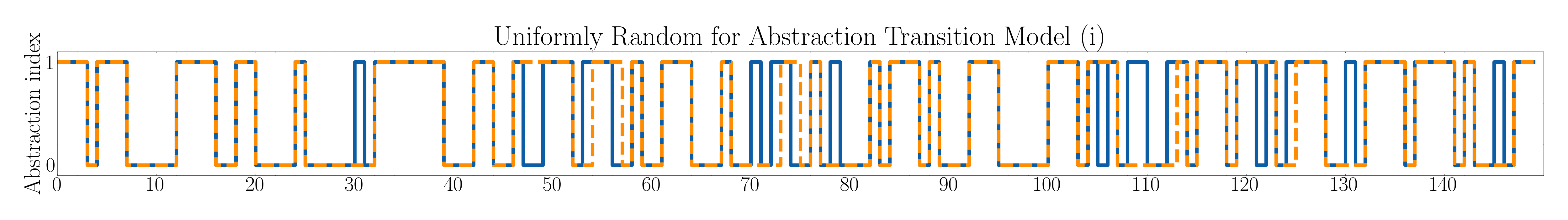} 
    \end{subfigure}
    \end{minipage}
    \vspace{-2.0mm}%
    
    \caption{%
    Visualizations of attention switching for the first 150 time steps of the different switching patterns using the first abstraction transition model. The true abstraction index is shown in solid blue while the inferred abstraction is shown as a dotted orange line. \textbf{Top:} Step occurs at time step 100. \textbf{Middle:} Periodic switching with a period of 10. \textbf{Bottom:} Uniformly random switching.%
    }
    \label{fig:example1_patterns}
    \vspace{-5.0mm}
\end{figure*}

{\bf Results.} Next we simulated different switching patterns for the true underlying contexts including \emph{step} (where the context remains constant for a number of time steps and then switches to the other context), \emph{periodic} (where the context switches back and forth periodically), and \emph{uniformly random} (where a context is chosen randomly at every time step) for 500 time steps and allowed the robot to randomly choose its actions. We also ran  these simulations for two different abstraction transition models where (i) the probability of maintaining the same abstraction is high ($p(\phi^i_t|\phi^i_{t-1}) = 0.8$), and (ii) the probability of maintaining the same abstraction or changing is uniform ($p(\phi^i_t|\phi^i_{t-1}) = 0.5$). Additionally, we provided the robot with noisy states under the same patterns. Specifically, we assign a probability, $\sigma$, that the observed state is another possible state rather than the true state.  Then we averaged across five seeds the accuracy,\footnote{Accuracy is calculated as the number of time steps the robot correctly inferred the abstraction over the time horizon.} average lag,\footnote{Lag is calculated as the number of time steps it takes for OAS to infer the correct abstraction after a context switch and before the next. Average lag is over the number of switches in the trial.} and maximum lag for the maximum likelihood abstraction determined by OAS.  These results are summarized in Table~\ref{tab:example1}. Example trials for a step, periodic, and random pattern for the first abstraction transition model are illustrated in Fig.~\ref{fig:example1_patterns}.

\begin{table}[h]

\vspace{1mm}
 \caption{Key statistics for the discrete-space tracking example}
\begin{center}

\begin{tabular}{ |p{2.3cm}|p{1.5cm}|p{1.5cm}|p{1.5cm}|  }
 \hline
\textbf{Switching Pattern} & \textbf{Accuracy} &\textbf{Avg. Lag} & \textbf{Max. Lag}\\
 \hline
 \hline
 \multicolumn{4}{|c|}{Abstraction Transition Model (i)} \\
 \hline
 \hline
 Step (at $t=100$)   & $0.99 \pm 0.00$ & $0.10 \pm 0.20$ & $0.20 \pm 0.40$\\
 \hline
 Noisy step (at $t=100, \sigma=0.7$)&  $0.99 \pm 0.00$  & $0.70 \pm 0.98$ & $0.80 \pm 1.17$ \\
 \hline
 Periodic (period = 10) &  $0.93 \pm 0.001$  &  $0.70 \pm 0.1$  & $6.80 \pm 1.72$\\
 \hline
 Noisy periodic (period = 10 , $\sigma=0.7$)&  $0.95 \pm 0.01$  & $0.48 \pm 0.05$ & $4.00 \pm 0.63$\\
 \hline
 Uniformly random & $0.84 \pm 0.01$ & $0.31 \pm 0.03$ & $3.80 \pm 1.17$ \\
 \hline
 \hline
 \multicolumn{4}{|c|}{Abstraction Transition Model (ii)} \\
 \hline
 \hline
 Step (at $t=100$)   &  $0.83 \pm 0.01$  & $0.00 \pm 0.00$ &  $0.00 \pm 0.00$ \\
 \hline
 Noisy step (at $t=100, \sigma=0.7$)& $0.83 \pm 0.01$   & $0.50 \pm 0.32$ & $1.00 \pm 0.63$\\
 \hline
 Periodic (period = 10 ) &  $0.84 \pm 0.01$   &  $0.22 \pm 0.06$  & $2.60 \pm 0.49$\\
 \hline
 Noisy periodic (period = 10, $\sigma=0.7$)&   $0.83 \pm 0.02$ & $0.24 \pm 0.11$  & $2.40 \pm 1.02$\\
 \hline
 Uniformly random & $0.84 \pm 0.01$ & $0.19 \pm 0.03$ & $2.40 \pm 0.49$ \\
 \hline

\end{tabular}

 \label{tab:example1}
\vspace{1mm}
\end{center}
\end{table}
\vspace{-5mm}
In general, the high accuracy values and low average lags for the first abstraction transition model demonstrate that OAS captured the attention switches well and quickly, with an average lag of 0.5 even for a challenging noisy periodic scenario. The expected robustness benefits of using attention are also demonstrated by the fact that there is virtually no change in performance on these statistics when a high state estimation noise value, $\sigma$, is introduced. The second abstraction transition model also performed well; however it experienced more bouncing behavior as seen in the step results where the lag is 0 (i.e., switched at the correct time) but lower accuracy than the first transition model. 

\subsection{Continuous-Space Tracking Example}

{\bf Scenario.} Here, we implement the scenario where a robot dog needs to pay attention to one of the two humans with the ``treat" in order to receive it (Fig. \ref{fig:hallway_setup}). In this scenario, the humans may walk around arbitrarily as long as the human with the treat stays within the robot's field of view. Meanwhile, the other human may move out of the field of view of the robot. Additionally, the humans may hand off the treat to the other human at any time. As a result, the robot needs to switch its attention to the new human in order to continue receiving treats. 


{\bf Hardware implementation.} For hardware implementation (Fig.~\ref{fig:hallway_setup}), we use the Unitree Go1 quadruped robot with an Intel Realsense RGB-D camera. The camera provides noisy observations of the state, $s_t = [x_1,y_1,x_2, y_2]$, corresponding to the positions of \texttt{human1} and \texttt{human2} relative to the robot. 
We utilize a pre-trained neural network (implemented as part of the TensorFlow Object Detection API \cite{Huang17}) in order to detect humans using RGB images; the depth image is then used to estimate the relative locations of the humans.
We also assume the ability to identify each human individually (since the state $s_t$ corresponds to the locations of each individual human); in our hardware setup, we implement this by distinguishing shirt colors.
The actions are constant velocity commands corresponding to left, right, straight, right and straight, and left and straight. Additionally the robot may choose to stop. The ``treat" corresponds to an AprilTag \cite{Olson11} that may be passed from human to human. 

{\bf Contexts.} Different contexts in this scenario arise from the reward function changing over time. Specifically, the reward changes from 1 to 0 when the robot is following a particular human who then passes the treat to the other human.
We highlight that this corresponds to a change in the reward \emph{function}, since the same state is assigned different reward values after the treat is passed.
We note that the robot does not utilize the AprilTag treat for the purpose of tracking; the AprilTag is simply used to assign rewards (specifically, a reward of 1 if the robot is close enough to the human that is holding the treat). 
Thus, the attention switch that the robot must perform is intuitively similar to a dog's: if it stops receiving treats, this serves as a signal to pay attention to the other human and follow it instead. 

\begin{figure}[t]
\centering
\begin{subfigure}{}
    \centering
    \includegraphics[width=0.2\textwidth]{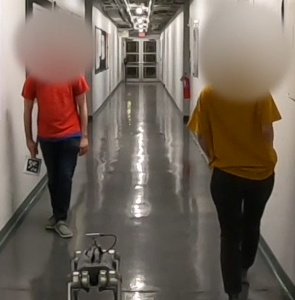}
\end{subfigure}
\begin{subfigure}{}
    \centering
    \includegraphics[width=0.2\textwidth]{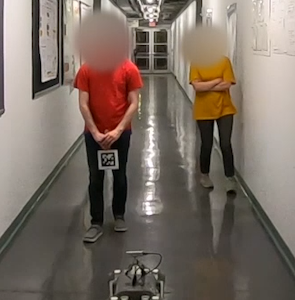}
\end{subfigure} \\
\vspace{-2mm}
\begin{subfigure}{}
    \centering
    \includegraphics[width=0.2\textwidth]{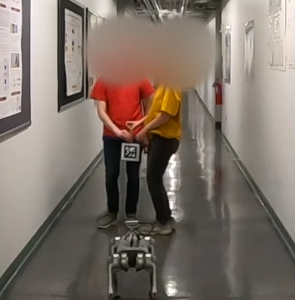}
\end{subfigure}
\begin{subfigure}{}
    \centering
    \includegraphics[width=0.2\textwidth]{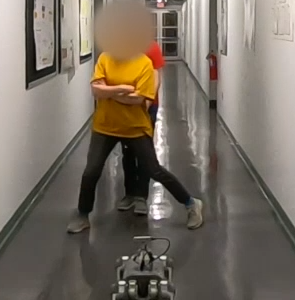}
\end{subfigure} \\
\vspace{-2mm}
\begin{subfigure}{}
    \centering
    \includegraphics[width=0.2\textwidth]{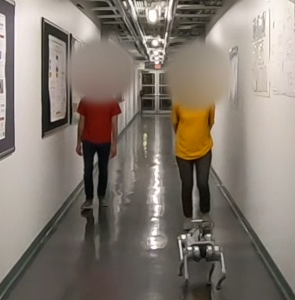}
\end{subfigure}
\begin{subfigure}{}
    \centering
    \includegraphics[width=0.2\textwidth]{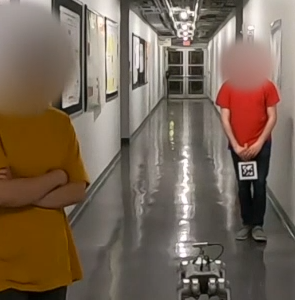}
\end{subfigure} \\
\vspace{-2mm}
    \begin{center}
     \footnotesize(a) \hspace{33mm} (b)
    \end{center}
\vspace{-3mm}
\caption{Instances of the continuous-space tracking example. \textbf{(a)} An example of a treat hand-off (AprilTag) between the two humans followed by an attention switch for the robot dog. Top: the robot dog is following \texttt{human1} (red) that has the treat. Middle: \texttt{human1} passes the treat to \texttt{human2} (yellow). Bottom: the robot dog has switched its attention correctly to \texttt{human2} and is tracking \texttt{human2} (not the treat). \textbf{(b)} An example of the robot dog ignoring the human that does not have the treat (\texttt{human2}). The progression of images (top to bottom) shows \texttt{human2} crossing from right to left in front of \texttt{human1} while the robot dog continues to follow \texttt{human1}.}
\label{fig:experiment}
\end{figure}

{\bf Abstractions.} Bisimulation on the two different contexts generates abstractions that capture the aforementioned intuition. In the context where \texttt{human1} has the treat, the state abstraction considers states $[x_1, y_1, x_2, y_2]$ and $[x_1, y_1, x_2', y_2']$ to be equivalent. In other words, the state abstraction focuses attention on the location of \texttt{human1} and ignores the location of \texttt{human2}. The robot's control policy is a function of this abstract state. By focusing only on context-relevant information (i.e., the location of the human who has the treat), the robot's policy is robust to noise in the state estimate of the human who does not have the treat. Thus, when the robot is paying attention to and following \texttt{human1}, the other human may wander far from the robot (leading to a noisy state estimate) without impacting the robot's performance. See the Appendix for more discussion on the sensor noise experienced for this example.

\textbf{Results.} For the trials, we had two humans, \texttt{human1} wearing red and \texttt{human2} wearing yellow, walk with the robot down a hallway. When not holding the treat, the human performed behaviors such as walking in and out of the robot's field of view, zig-zagging behind the human with the treat, and crossing in front of the human with the treat. Meanwhile, the human holding the treat often blocked the treat from view when moving around in order to highlight that the robot was paying attention to and tracking them (rather than the AprilTag). Additionally, each trial contained 1-3 treat hand-offs to the other human and lasted approximately 1 minute (with the robot operating at 5 Hz). The abstraction transition model set the probability of maintaining the same abstraction as 0.8. We refer the reader to our video\footnote{See video of results \url{https://youtu.be/lcL1Pcf3qmk}.} to view these behaviors, as well as qualitatively view the robot inferring context switches using the OAS algorithm and switching its attention and behavior accordingly. Two example interactions are shown with picture progressions in Fig. \ref{fig:experiment}.
\begin{table}[h]
 \caption{Key statistics for the continuous-space tracking example}
\begin{center}

\begin{tabular}{ |p{2.3cm}|p{1.5cm}|p{1.5cm}|p{1.5cm}|  }
 \hline
\textbf{Accuracy} & \textbf{Average Lag (sec)} &\textbf{Max. Lag (sec)} & \textbf{Normalized Reward}\\
 \hline
 $0.91 \pm 0.07$ & $0.37 \pm 0.24$    & $2.31 \pm 1.48$ & $0.27 \pm 0.11$ \\
 \hline

\end{tabular}

 \label{tab:example2}

\end{center}
\end{table}
\vspace{-4mm}

We summarize key statistics (introduced in Sec.~\ref{sec:discrete example}) of our OAS approach averaged across 12 trials in Table~\ref{tab:example2}. In general, the high accuracy indicates that the robot performed well at maintaining its attention on the correct human. The normalized reward is the cumulative reward averaged over the time horizon of a trial. Note that the humans could decide to act adversarially, e.g., walking further away as the robot approached them. The lags (reported in seconds where one time step corresponds to $0.2$s each) were very low, indicating that the attention switches occurred in a timely manner.

\section{Conclusions and Future Work}

We have presented an approach for allowing a robot to determine what context-relevant information it should attend to, infer when the context (i.e., task or dynamics) has switched (among a catalog of potential contexts), and switch its attention accordingly. To achieve this, we leverage bisimulation-based state abstraction techniques in order to define an attention mechanism in a given context. A Bayesian inference algorithm then operates online in order to detect context switches by maintaining a belief that is updated using reward and state observations. We demonstrated our method on an illustrative discrete-state tracking example and hardware demonstrations of a tracking problem using a quadruped robot. 

\textbf{Challenges and Future Work.} There are a number of exciting directions for future work. First, we expect that our approach will generalize to state abstraction techniques beyond the (strict) notion of bisimulation used here (e.g., to metric-based bisimulation notions). Second, it may be of practical interest to relax the finiteness assumption on the catalog of possible state abstractions, and allow for a continuously parameterized set by using Bayesian inference techniques that operate on continuous spaces (e.g., particle filters). Finally, a particularly interesting direction is to synthesize policies that allow the robot to \emph{actively} minimize uncertainty in its belief about the appropriate state abstraction to utilize. This potentially requires a careful tradeoff between actions that reduce uncertainty with those that make progress towards the task.






\section*{Acknowledgements}

The authors were supported by the School of Engineering
and Applied Science at Princeton University through the generosity of William Addy ’82, the NSF CAREER award [2044149], and the Office of Naval Research [N00014-21-1-2803]. The authors are also very thankful for the input provided by Eric Lepowsky and Susan Redmond.

\appendix

This appendix provides additional details for the depth sensor noise experienced in the continuous-space tracking example in Sec. \ref{sec:examples}.

\begin{figure}[h]
\centering
\includegraphics[width=0.45\textwidth]{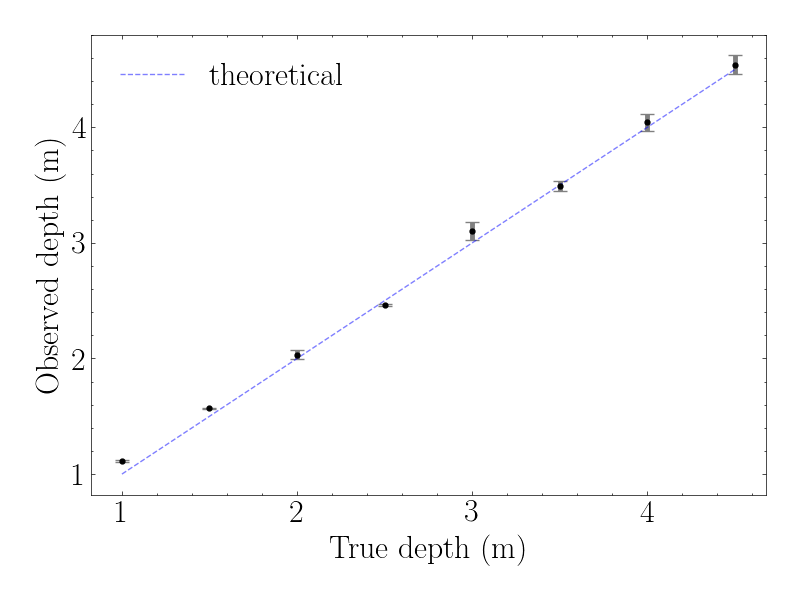}
\vspace{-2mm}
\caption{Depth measurements taken from the Intel Realsense RGB-D camera in the test hallway compared to the true depth of a human. Averages were obtained from 10 measurements. Notice that the sensor noise (standard deviation) increases with distance.}
\label{fig:sensor}
\end{figure}

We illustrate the state noise from the camera we use in Fig. \ref{fig:sensor}. We use an Intel Realsense RGB-D camera and measured the depth of an observed human in the test hallway. We note that this noise scales with the distance from the robot.

\newpage
\bibliographystyle{IEEEtran}
\bibliography{main}


\end{document}